\documentclass[conference]{IEEEtran}
\usepackage{times}

\usepackage[numbers]{natbib}
\usepackage{multicol}
\usepackage[bookmarks=true]{hyperref}
\usepackage[pdftex]{graphicx}
\DeclareGraphicsExtensions{.pdf,.jpeg,.png}
\usepackage{soul,xcolor}
\setstcolor{red}
\usepackage{hyperref}
\hypersetup{colorlinks,allcolors=red}
\begin{document}

\title{Data-Link: High Fidelity Manufacturing Datasets for Model2Real Transfer under Industrial Settings}

\author{Sunny Katyara, Mohammad Mujtahid, Court Edmondson}



%

\maketitle

\begin{abstract}
High-fidelity datasets play a pivotal role in imbuing simulators with realism, enabling the benchmarking of various state-of-the-art deep inference models. These models are particularly instrumental in tasks such as semantic segmentation, classification, and localization. This study showcases the efficacy of a customized manufacturing dataset comprising 60 classes in the creation of a high-fidelity digital twin of a robotic manipulation environment. By leveraging the concept of transfer learning, different 6D pose estimation models are trained within the simulated environment using domain randomization and subsequently tested on real-world objects to assess domain adaptation. To ascertain the effectiveness and realism of the created data-set, pose accuracy and mean absolute error (MAE) metrics are reported to quantify the model2real gap.
\end{abstract}

\IEEEpeerreviewmaketitle

\section{Introduction}
We are living in a world where vast amounts of data hold the power to revolutionize industries and shape the future of manufacturing. Datasets play a pivotal role in virtually every field within today's digital world, enabling data-driven decision-making. In the manufacturing industry, data-sets assume a critical position, offering invaluable insights to enhance product quality, optimize production processes, streamline supply chains, and achieve heightened operational efficiencies. However, creation of data-sets represents a laborious and time-consuming endeavour, necessitating the acquisition of high-quality, consistent, scalable, and adaptable data for process automation and optimization, particularly in the realms of robotic grasping and manipulation, assumes utmost significance.

The process of annotating 6D poses in data-sets for robotic grasping and manipulation represents a labour-intensive endeavour, surpassing the challenges encountered in 2D image labelling. To mitigate this challenge, a viable solution lies in the utilization of synthetic data, which offers meticulously annotated samples at a low cost for training pose estimation models \cite{c1}\cite{c2}. However, the substantial disparities between synthetic (source) and real (target) models result in sub-optimal performance. To bridge this gap, a promising approach emerges, combining domain randomization and photo-realistic synthetic data \cite{c3}\cite{c4}, aiming to address the domain shift between the source and target domains. In our study, we adopt the real-sim-real transfer method as a means of domain adaptation to overcome sensor noise and realism issues.

\begin{figure}[t]
      \centering
      \includegraphics[width=9 cm]{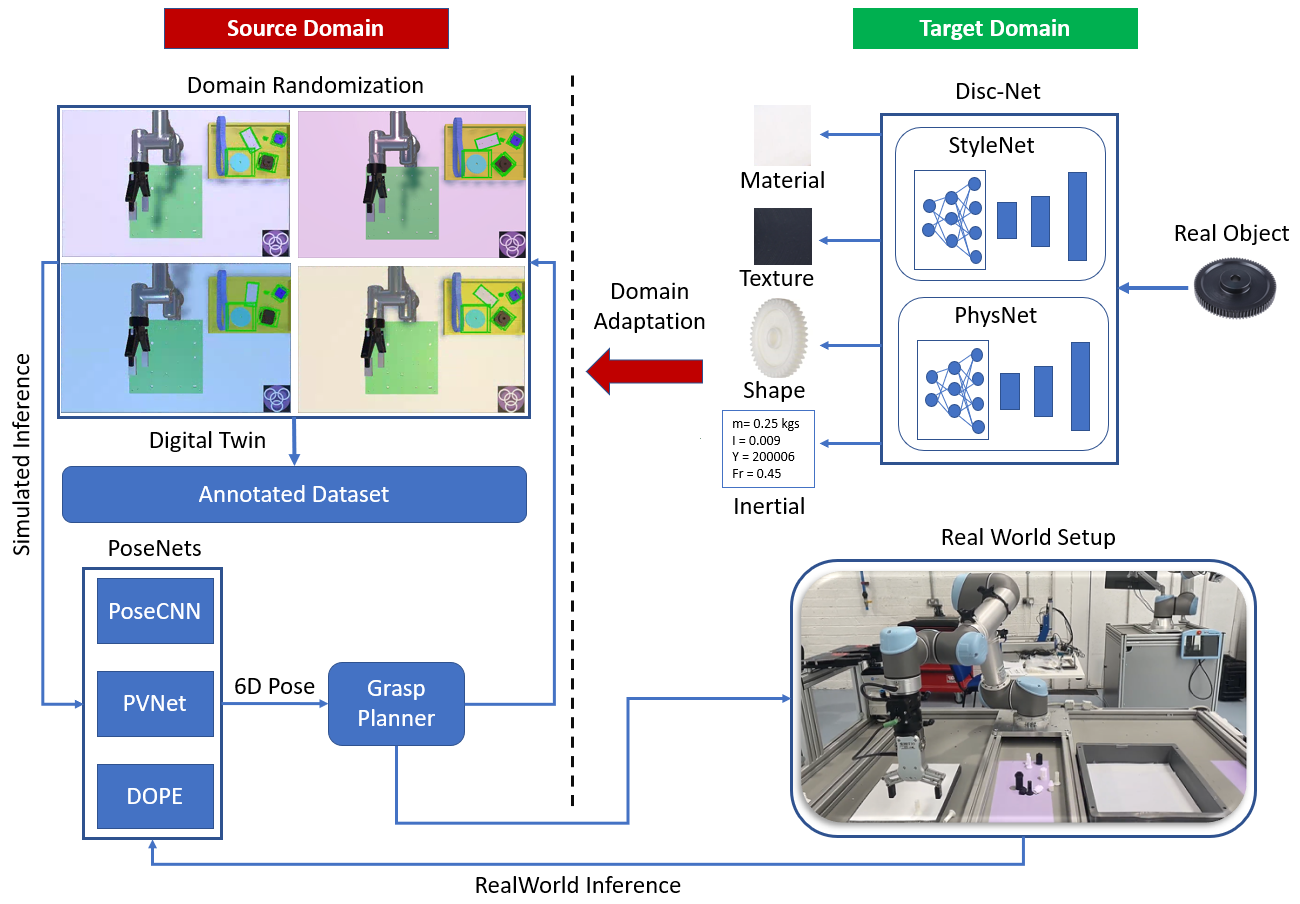}
      \caption{Data-Link pipeline for creating a high-fidelity dataset to bridge the domain gap for moder2real transfer under industrial conditions.}
      \label{one}
      \vspace{-15pt}
   \end{figure}

While certain generic data-sets, such as YCB Videos \cite{c5}, MVTech AD \cite{c6}, and Dex-Net 2.0 \cite{c7}, have been employed for training models in semantic segmentation, classification, and localization, their limitations become apparent in terms of restricted object variety and the absence of real-world manufacturing context. Consequently, our research proposes the creation of an extensive, high-fidelity data-set encompassing a range of 3D objects commonly employed within the manufacturing industry. As illustrated in Fig. \ref{one}, we discretize the captured 3D object, acquired using a high-resolution camera, into descriptive components, comprising texture, material, shape, and inertial dynamics. This process is facilitated by a customized neural network known as Disc-Net, which accepts RGBD data and CAD models of the object of interest as inputs, enabling the extraction of desired object features. The proposed Disc-Net architecture incorporates two distinct neural network components: style extraction, encompassing texture and material, while the other focuses on shape and inertial parameters. These extracted features are subsequently transferred to synthetic models within the digital twin environment, where they are annotated using a standard bounding box annotator \cite{c8}. The resulting synthetic annotated data-set is then utilized for training various pose estimation networks including PoseCNN \cite{c5}, PVNet \cite{c9}, and DOPE \cite{c10} and assessing their performance within real-world settings, thereby providing a benchmark for evaluating the realism and effectiveness of generated data-set for sim2real transfer.

In summary, this research makes following contributions:

\begin{itemize}
    \item We propose a robust pipeline that facilitates extraction of desired rendering and physics features from real objects, enabling their seamless transfer to synthetic models. This approach allows for low-cost augmentation and creation of a synthetic data-set with comprehensive 6D annotations, operating under the paradigm of domain adaptation.
    \item •	Use high-fidelity synthetic dataset to train different PoseNets within designed digital twin under domain randomization and eventually evaluate their performance on real world setup using pose accuracy and mean absolute error (MAE) metrics to quantify domain gap. 
    
\end{itemize}

\begin{figure}[t]
      \centering
      \includegraphics[width=8.5 cm]{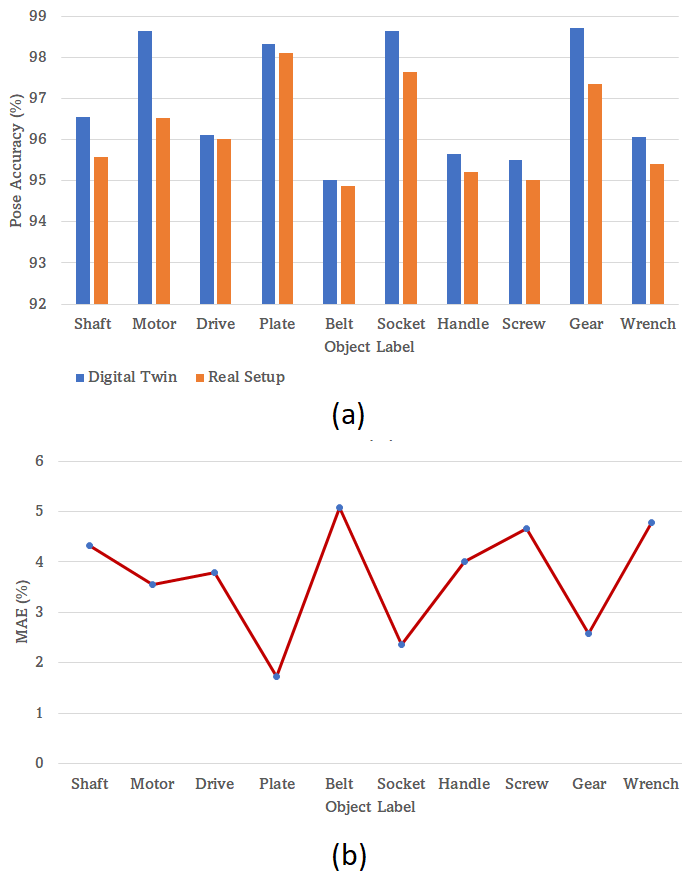}
      \caption{Performance evaluation of PoseNets trained using synthetic data and tested on real-world object instances (a) Pose accuracy is higher than 90\%, (b) MAE is less than 5\% threshold.}
      \label{two}
      \vspace{-15pt}
   \end{figure}

\section{Methodology and Discussion}

The primary objective of this research is to leverage 6D pose estimation models trained on low-cost synthetic data for real-world settings, eliminating the need for laborious fine-tuning or expensive model retraining. The proposed methodology encompasses the extraction of style features, including texture and material properties, using StyleNet. StyleNet is an autoencoder architecture with an additional layer dedicated to semantic understanding of texture and material properties derived from latent space representation. In addition, shape and inertial parameters of 3D objects of interest are extracted using PhysNet, which combines a modified PointNet \cite{c11} with a customized NeRF architecture \cite{c12}. These two networks are integrated into a composite model architecture named Disc-Net, enabling simultaneous inference. The StyleNet layer takes RGB data as input, while the PhysNet layer processes PointCloud data and CAD models of candidate objects, yielding descriptive descriptors encompassing material, texture, shape, and inertial properties from the composite model (DiscNet). 

Using the extracted descriptors from the real world, virtual objects are developed within the Unity 3D engine to establish a digital twin environment for manipulation scenarios, serving as a domain adaptation technique with a specific focus on bin-picking tasks. To ensure clarity and comprehensiveness, we have modelled 60 distinct manufacturing object classes, ranging from motors and gears to wrenches and Allen keys. These models form the basis for generating a synthetic 6D annotated dataset, incorporating domain randomizations. The randomizations encompass variations not only in rendering-related factors, such as lighting, object positions, and camera locations, but also in physics parameters, including mass, friction, inertias, and other relevant factors. The dataset consists of 3000 training and 1800 test images, each with dimensions of 640 $\times$ 480 pixels. These images are utilized to train three distinct PoseNets: PoseCNN, PVNet, and DOPE. Operating on RGB images as input, these PoseNets estimate 6D poses of objects of interest within the scene, leveraging 2D to 3D correspondences. The estimated poses from these models are subsequently utilized by the Moveit API for grasp planning, with evaluation conducted in both virtual and real setups.

The experimental findings indicate that despite being trained on synthetic data, all the networks exhibit comparable performance levels during inference for various objects within both the digital twin and real setups. This observation is substantiated by the outcomes presented in Fig. \ref{two}, illustrating the pose accuracy and MAE values. However, it is important to note that the grasp planner requires fine-tuning when adapting to real setups. This necessity arises due to unmodeled dynamics, including object slip, external disturbances, stochasticity, and inherent limitations in the planner's adaptability.

\section{Conclusion} 
\label{sec:conclusion}

This research proposed a novel methodology for developing high-fidelity manufacturing data-set using domain adaptation and domain randomization strategies. The methodology leveraged digital twins of manipulation scenarios to eliminate the need for laborious manual 6D annotations. Our approach encompassed not only the extraction of style features but also the incorporation of inertial dynamics, aiming to achieve a heightened level of realism and model plausibility in rendering and object interactions. The evaluation results demonstrated that our method effectively bridges the domain gap, as evidenced by lower values of MAE and higher pose accuracy rates over sim2real tests. Importantly findings indicated that the proposed approach not only demonstrated superior performance for larger objects but also exhibits notable adaptability to smaller components, even when employing low-cost camera sensors.

This research, however, addresses the reality gap in perceptual pipelines by enabling them to adapt to diverse environments without the need for fine-tuning. Nonetheless, it is important to note that with domain changes, the motion planners also require tuning. In future endeavors, our objective is to extend this work by developing a control pipeline that harnesses the capabilities of reinforcement learning algorithms. The overarching goal is to devise manipulation sequences that are both robust and optimal, thereby facilitating enhanced adaptability and efficiency in manipulation tasks \cite{c13}.

\section*{Acknowledgments}

Authors would like to extend their sincere gratitude to the Industrial Steering Board (ISB) at IMR, Ireland, for generously providing the funding required to undertake this significant research endeavor. This study aims to address the gap between AI-driven academic research outcomes and the prevailing traditional industrial issues.

\end{document}